\pdfoutput=1

\documentclass[11pt]{article}

\usepackage[final]{acl}

\usepackage{times}
\usepackage{latexsym}

\usepackage{algorithm}
\usepackage{algorithmic}
\usepackage{xcolor}
\usepackage{makecell}
\usepackage{multirow}
\usepackage{enumitem}
\usepackage[normalem]{ulem}
\usepackage{hwemoji}
\usepackage{pifont} 

\usepackage{booktabs}
\usepackage{adjustbox}

\usepackage[T1]{fontenc}

\usepackage[utf8]{inputenc}

\usepackage{microtype}

\usepackage{inconsolata}

\usepackage{graphicx}

\title{\textsc{CoBA}: Counterbias Text Augmentation \\ for Mitigating Various Spurious Correlations via Semantic Triples}

\author{
  Kyohoon Jin\textsuperscript{1,4}* \quad
  Juhwan Choi\textsuperscript{2,5}* \quad
  Jungmin Yun\textsuperscript{5} \quad \\ 
  {\bf Junho Lee\textsuperscript{3}} \quad
  {\bf Soojin Jang\textsuperscript{4}} \quad
  {\bf Youngbin Kim\textsuperscript{4,5 \dag}}\thanks{\ \ * denotes euqal contribution. \\ \dag Corresponding author. Email: \texttt{ybkim85@cau.ac.kr}}
  \\
  \textsuperscript{1}DATUMO \quad
  \textsuperscript{2}AITRICS \quad
  \textsuperscript{3}Brainventures \quad \\
  \textsuperscript{4}Graduate School of Advanced Imaging Sciences, Multimedia and Film, Chung-Ang University \\
  \textsuperscript{5}Department of Artificial Intelligence, Chung-Ang University
  \\
  \texttt{kyohoon.jin@selectstar.ai}, \texttt{jhchoi@aitrics.com}
  \texttt{panic@brainventur.com } \\
  \texttt{\{cocoro357, sujin0110, ybkim85\}@cau.ac.kr}
  \\
}

\begin{document}
\maketitle
\begin{abstract}
Deep learning models often learn and exploit spurious correlations in training data, using these non-target features to inform their predictions. Such reliance leads to performance degradation and poor generalization on unseen data. To address these limitations, we introduce a more general form of counterfactual data augmentation, termed \textit{counterbias} data augmentation, which simultaneously tackles multiple biases (e.g., gender bias, simplicity bias) and enhances out-of-distribution robustness. We present \textsc{CoBA}: \textbf{Co}unter\textbf{B}ias \textbf{A}ugmentation, a unified framework that operates at the semantic triple level: first decomposing text into subject-predicate-object triples, then selectively modifying these triples to disrupt spurious correlations. By reconstructing the text from these adjusted triples, \textsc{CoBA} generates \textit{counterbias} data that mitigates spurious patterns. Through extensive experiments, we demonstrate that \textsc{CoBA} not only improves downstream task performance, but also effectively reduces biases and strengthens out-of-distribution resilience, offering a versatile and robust solution to the challenges posed by spurious correlations.
\end{abstract}

\section{Introduction}

Despite deep learning’s success across various domains, spurious correlations continue to pose significant challenges in training effective models \cite{ye2024spurious}. Spurious correlations are patterns that appear in datasets but do not represent genuine relationships, such as correlations with background or textures \cite{beery2018recognition, geirhos2019imagenet, sagawa2020distributionally}. This phenomenon is also prevalent in text data, where spurious correlations frequently emerge at the word-level. In such cases, certain words or phrases become associated with specific labels due to their co-occurrence in particular contexts. This association often fails to reflect the actual meaning or intent, resulting in performance degradation in models \cite{wang2022identifying, joshi2022all, chew2024understanding}. Furthermore, spurious correlations are linked to various biases, including gender bias, and challenges related to out-of-distribution (OOD) robustness. Consequently, mitigating these correlations is crucial for enhancing deep learning models in a broader context \cite{mcmilin2022selection, liusie2022analyzing, ming2022impact, zhou2024explore}.

Early studies have focused on mitigating spurious correlations from a model-centric perspective by identifying spurious features. For instance, several approaches suggested reweighting data samples to mitigate spurious features; however, this strategy can inadvertently introduce new biases by overemphasizing irrelevant features~\cite{han2021influence, shi2023re}. Subsequently, recent studies have shifted the focus toward a data-centric approach, particularly in the field of natural language processing~\cite{ye2024spurious}. In this direction, researchers have been exploring data manipulation techniques aimed at enhancing the generality and diversity of data distribution and reducing the impact of spurious correlations present in the original data, thereby improving model capabilities~\cite{wang2024survey}. A line of studies suggest that augmenting datasets with counterfactual data—entailing minimal modifications to the original sentences—can effectively mitigate spurious correlations~\cite{kaushik2020learning, udomcharoenchaikit2022mitigating, chan2023spurious}. While early studies relied on human-annotated counterfactuals, more recent works propose automatically generating them through data augmentation, demonstrating their effectiveness in reducing spurious correlations \cite{zeng2020counterfactual, wang2021robustness, wen2022autocad, treviso2023crest, sachdeva2024catfood}. However, due to the minimal modifications, this approach may lack diversity, potentially leading to issues such as overfitting and subsequent performance degradation~\cite{qiu2024paircfr}.

In this study, we extend current research on counterfactual data augmentation to \textit{counterbias data augmentation}, which simultaneously addresses various biases and challenges, such as gender bias, and out-of-distribution robustness. Although counterfactual data has been effective in mitigating spurious correlations, there remains significant potential for a unified approach that can concurrently tackle these diverse challenges. To explore this, we propose transforming the given text into a set of semantic triples using a large language model (LLM), with each triple encapsulating compressed information from the sentences. By generating counterfactual triples through modifications of the original triples and reconstructing text from these debiased triples using an LLM, we can create augmented \textit{counterbias} data. This triple-level modification simplifies the generation of counterfactuals, as triples naturally contain the key elements of sentences. Additionally, with the support of LLMs in reconstructing text from triples, our framework can effectively diversify augmented text. \textit{Counterbias data augmentation} differs from previous counterfactual data augmentation approaches, which aim to make minimal changes while flipping the original data’s label.

Additionally, we conducted an analysis to identify principal words in various models using word importance measurements,  revealing that each model has a distinct set of principal words. This finding suggests that counterbias data generated for a single model may not be effective for other models. To address this finding, we employ a majority-voting-based ensemble method to identify words that may contribute to spurious correlations. This approach is effective for augmenting counterbias data that can be universally applied across various models. Through experiments validating the effectiveness of our proposed framework, \textsc{CoBA}, we observed that it effectively alleviates various biases and challenges while also augmenting counterbias data applicable across different models.

Our main contributions are as follows: 

\begin{itemize}[leftmargin=*]
\item \textbf{A Unified Framework for Counterbias Augmentation:} We introduce \textsc{CoBA}, a novel approach that extends counterfactual data augmentation to \textit{counterbias} data augmentation. Unlike prior methods that primarily focus on minimal label-flipping modifications, \textsc{CoBA} targets a broader range of biases and spurious correlations, improving both in-distribution performance and out-of-distribution robustness. 

\item \textbf{Insights into Spurious Correlations Across Models:} Through a detailed analysis of word importance, we reveal how spurious correlations vary significantly across different model architectures, underscoring the limitations of relying on a single model. This insight motivates our ensemble-based strategy to identify spurious correlations more reliably. 

\item \textbf{Empirical Validation and Practical Benefits:} Extensive experiments across tasks like sentiment analysis, natural language inference, and text style transfer show that \textsc{CoBA} consistently alleviates multiple biases and enhances model resilience to distribution shifts. These results highlight \textsc{CoBA}’s versatility and its potential to construct more robust, fair, and generalizable deep learning models. \end{itemize}

\section{Related Work}

Counterfactual data augmentation has been shown to mitigate spurious correlations effectively. An early study introduced the concept of counterfactual data by manipulating existing data to alter the label with minimal modifications \cite{kaushik2020learning}. These counterfactual data have been demonstrated to be useful for mitigating spurious patterns and evaluating local decision boundaries of models \cite{gardner2020evaluating}.

Since these studies relied on human annotators to generate counterfactual data, producing such data for various datasets was challenging. As a result, researchers began exploring automated methods for generating counterfactual data, particularly in data augmentation setups. In early explorations, predefined rules were applied to augment counterfactual data \cite{zmigrod2019counterfactual, wang2021robustness}.

Beyond rule-based techniques, deep learning models have been employed to augment counterfactual data. For example, several studies have proposed leveraging well-trained classifiers to identify principal words  \cite{wang2022identifying, wen2022autocad, bhan2023tigtec}. Additionally, generating counterfactual data using deep learning models has been proven to be effective in diversifying the generated data \cite{wu2021polyjuice, treviso2023crest, sun2024acamda, ding2024rationale}. Recently, researchers have also begun exploring the use of LLMs for counterfactual data augmentation \cite{sachdeva2024catfood, chang2024counterfactual, li2024prompting}. While C2L similarly adopts a collective decision strategy over counterfactuals, it relies on a single model; in contrast, CoBA leverages multiple diverse models to more robustly identify causal features and mitigate spurious correlations~\cite{choi2022c2l}.

\section{Methodology}

\begin{figure*}[t]
    \centerline{
    \includegraphics[width=0.97\textwidth]{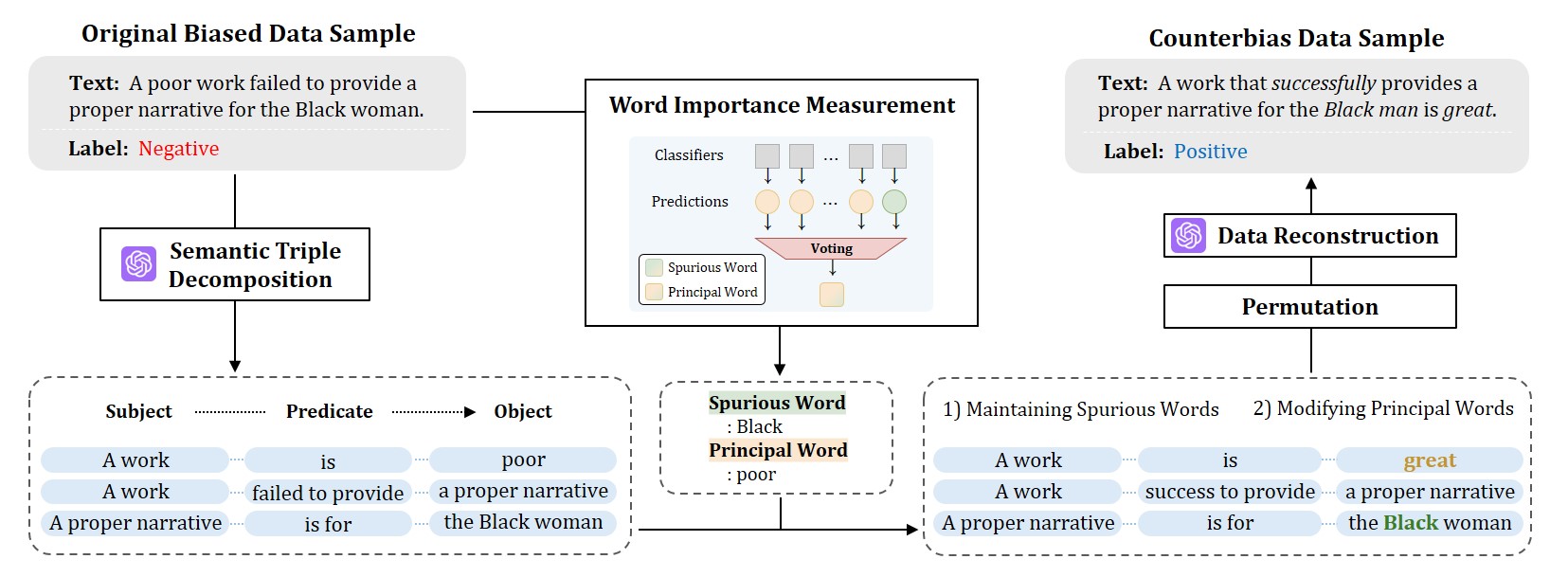}}
    \caption{Overall procedure of \textsc{CoBA}.}
\label{fig:framework}
\end{figure*}

\subsection{Overview}
In this paper, we aim to alleviate various biases and obstacles by mitigating spurious correlations through counterbias data augmentation. Specifically, given an original dataset $\mathcal{D}_{\textit{\small{ori}}}=\{(x_i, y_i)\}^{|\mathcal{D}_{\textit{\small{ori}}}|}_{i=1}$, where $x_i$ denotes the input text and $y_i$ its corresponding label, our goal is to generate new pairs $(\hat{x}_i, \hat{y}_i)$, where $\hat{x}_i$ is the augmented counterbias text and $\hat{y}_i \neq y_i$. We define counterbias text as text that retains the spurious words and semantics of the original text but is assigned a different label to mitigate spurious correlations, similar to counterfactual text. This definition represents a relaxed form of counterfactual text, which refers to text with minimal differences from the original data but with different labels \cite{molnar2020interpretable}. Unlike counterfactual data, counterbias data are not restricted to minimal changes; they may exhibit different syntactic structures and expressions, as long as they preserve the spurious words and semantics of the original text. This distinction allows counterbias data to introduce a wider variety of patterns, thereby amplifying the augmentation effect.

To accomplish this, we first decompose $x_i$ into a set of semantic triples, denoted by $T^{(x_i)}=\{t^{(x_i)}_j\}^{|T^{(x_i)}|}_{j=1}$. Each triple $t^{(x_i)}_j$ consists of (subject, predicate, object) derived from a sentence in $x_i$\footnote{Note that the relationship between a triple and a sentence is not one-to-one; a sentence can contain multiple triples.}. This decomposition is carried out by an LLM using a designated prompt. Next, we modify each decomposed $t^{(x_i)}_j$ to mitigate spurious correlations at the triple level, resulting in modified triples $\hat{t}^{(x_i)}_j \in \hat{T}^{(x_i)}$. Specifically, we proceed as follows:

\textbf{1)} We identify two sets of words, $W_s$ and $W_p$, where $W_s$ is the set of spurious words that cause spurious correlations, and $W_p$ is the set of principal words crucial for determining the label of $x_i$. To obtain $w_s \in W_s$ and $w_p \in W_p$, we use multiple classifiers with different backbones trained on $\mathcal{D}_{\textit{\small{ori}}}$ and word importance measurement techniques.

\textbf{2)} We then generate $\hat{T}^{(x_i)}$ by modifying any $t^{(x_i)}_j$ that contains $w_p$, while retaining $t^{(x_i)}_j$ that contains $w_s$. This minimal, triple-level alteration differentiates the label yet preserves spurious words, producing what we term counterfactual triples.

\textbf{3)} Lastly, to introduce diverse patterns into the augmented data—thereby enhancing OOD robustness—we randomly permute the order of the triples in $\hat{T}^{(x_i)}$ and delete several triples as well.

Finally, we augment $\hat{x}_i$ by reconstructing counterbias text from the modified $\hat{T}^{(x_i)}$ using the LLM with a designated prompt. Since we have modified $T^{(x_i)}$ to $\hat{T}^{(x_i)}$ to retain $\hat{y}_i$ instead of $y_i$ by modifying $t^{(x_i)}_j$ with $w_p$, the reconstructed $\hat{x}_i$ contains the label $\hat{y}_i \neq y_i$. This results in a counterbias-augmented dataset $\mathcal{D}_{\textit{\small{cb}}}$, which is used to train a downstream task model by $\mathcal{D}_{\textit{\small{ori}}} \cup \mathcal{D}_{\textit{\small{cb}}}$. Figure~\ref{fig:framework} illustrates this overall procedure.

\begin{table}[t]
\centering
\resizebox{0.99\columnwidth}{!}{%
\begin{tabular}{c|ccc}
\Xhline{3\arrayrulewidth}
      & LIME  & IG & SV\\ \hline\hline
SST-2 & 26.72\% (83.9\%) & 18.53\% (83.2\%)                                                         & 14.21\% (85.3\%)                                                  \\ \hline
IMDB  & 8.64\% (81.4\%) & 7.00\% (74.6\%)                                                        & 7.99\% (81.3\%)                                                  \\ \Xhline{3\arrayrulewidth}
\end{tabular}%
 } 
\caption{The ratio of duplication among the top-5 most principal words for each model. The number in parentheses indicates the degree of overlap between two or more models, but not every model.}
\label{tab:word-importance}
\end{table}

\subsection{Analysis on Important Words}
\label{sec:method-analysis-important}
Before introducing \textsc{CoBA} in detail, we first present an analysis to investigate the differences in $W_p$ across various models from two perspectives. Implementation details are provided in Appendix~\ref{app:exp-setup-word-importance}.

\subsubsection{Word-level Importance Analysis}

First, we compared the word-level importance of each model. We measure word-level importance for each model using three different word importance measurement techniques: local interpretable model-agnostic explanations (LIME) \cite{ribeiro2016should}, integrated gradient (IG) \cite{sundararajan2017axiomatic}, and Shapley value (SV) \cite{rozemberczki2022shapley}.

We used these techniques to measure the top-5 important words for each model on the SST-2 and IMDB datasets. Afterward, we evaluated the ratio of duplicated words among the important words identified by each model. Specifically, we counted instances where all four models contained at least one duplicated word. Table~\ref{tab:word-importance} presents the results. The findings suggest that the number of words consistently regarded as important across all models is small. Notably, this ratio was less than 10\% of the total words in IMDB, which contains relatively longer text than SST-2. Although the models in this analysis share BERT-family architecture, they focus on different words when making predictions. However, when examining the overlap in important words between just two models at a time, we found that most cases exceeded 80\%. This indicates that while each model has its own tendencies, there is still meaningful overlap in the patterns they recognize, suggesting they focus on sentence semantics in distinct yet related ways.

\begin{table}[t]
\centering
\resizebox{\columnwidth}{!}{%
\begin{tabular}{c|cccc}
\Xhline{3\arrayrulewidth}
\textbf{SST-2}          & Noun          & Verb          & Adjective \& Adverb & Others        \\ \hline\hline
\texttt{BERT-base}     & \textbf{50.9} & 10.1          & 18.1                & \textit{20.9} \\
\texttt{BERT-large}    & \textbf{48.6} & 2.3           & \textit{35.1}       & 14.0          \\
\texttt{RoBERTa-large} & \textbf{67.9} & 9.8           & \textit{15.6}       & 6.7           \\
\texttt{BART-base}     & \textbf{35.3} & \textit{22.0} & 21.0                & 21.6          \\
\texttt{T5-base}     & \textit{25.5} & \textbf{40.1}          & 23.8                & 10.6 \\\Xhline{2\arrayrulewidth}
\textbf{IMDB}          & Noun          & Verb          & Adjective \& Adverb & Others        \\ \hline\hline
\texttt{BERT-base}     & \textit{22.6} & 12.2          & 16.0                & \textbf{49.2} \\
\texttt{BERT-large}    & \textit{28.4} & 12.4          & 8.3                 & \textbf{50.8} \\
\texttt{RoBERTa-large} & \textbf{46.7} & 16.0          & \textit{29.1}       & 8.2           \\
\texttt{BART-base}     & \textit{26.8} & 11.3          & 24.9                & \textbf{37.0} \\
\texttt{T5-base}     & 6.0 & 18.5          & \textit{24.4}                & \textbf{51.1} \\\Xhline{3\arrayrulewidth}
\end{tabular}
}
\caption{The ratio of POS tags among top-5 most important words for each model on SST-2 and IMDB. Bolded values represent the most frequent POS tag for each model and dataset, while italicized values represent the second most frequent POS tag.}
\label{tab:word-pos}
\end{table}

\subsubsection{POS Tagging Analysis} 
\label{sec:method-analysis-important-pos}

To support the findings of the previous analysis, we conducted an additional analysis by performing POS tagging on top-5 important words identified from the analysis above. Table~\ref{tab:word-pos} presents the result of this analysis. The findings indicate that the important words identified by each model have different POS tags, revealing that each model focuses on different aspects of the given text. Appendix \ref{app:exp-qualitative-pos} provides qualitative evaluation of this tendency.

These two analyses suggest that counterfactual data augmented by previous methods, which leveraged a single model to identify important words from the input text, may not be adequate for other models, diminishing the efficiency to be applied universally across various models. Inspired by this finding, we propose leveraging multiple models and using a majority-voting-based ensemble method to identify important words, including spurious and principal words.

\subsection{\textsc{CoBA}}
\label{sec:method-coba}

In this section, we introduce the detailed procedure of our \textsc{CoBA}; \textbf{Co}unter\textbf{B}ias \textbf{A}ugmentation framework. \textsc{CoBA} consists of three major components: semantic triples decomposition, triple-level manipulation, and reconstruction of counterbias text. The overall procedure is illustrated in Figure~\ref{fig:framework}.

\subsubsection{Semantic Triple Decomposition}

To augment the given $x_i$ into $\hat{x}_i$, we first decompose $x_i$ into a set of semantic triples $T^{(x_i)}=\{t^{(x_i)}_j\}^{|T^{(x_i)}|}_{j=1}$. Each $t^{(x_i)}_j$ represents a sentence or phrase from $x_i$, and follows the structure of (subject, predicate, object). For instance, given $x_i$ as ``I love In-N-Out. Their burger feels incredibly fresh'', the desired $T^{(x_i)}$ is \{(I, love, In-N-Out), (Their burger, feels, incredibly fresh)\}. While various techniques exist for triple decomposition, they primarily focus on decomposing a \textit{single} sentence into semantic triples, which differs from our purpose \cite{tan2019jointly, ye2021contrastive, chen2021semantic}.

To effectively decompose text containing multiple sentences into semantic triples, we utilize LLMs, which can perform various tasks when given proper instructions through prompts \cite{brown2020language, ouyang2022training}. We achieve this by designing a dedicated prompt $p_{\textit{\small{ext}}}$\footnote{Please refer to Appendix \textcolor{red}{\ref{app:prompt-design}} for full details on $p_{\textit{\small{ext}}}$ and other prompts.} for an LLM $\mathcal{L}$. Consequently, the desired set of semantic triples $T^{(x_i)}$ is obtained by $T^{(x_i)} = \mathcal{L}(x_i, p_{\textit{\small{ext}}})$.

\subsubsection{Triple-level Manipulation}

Since $T^{(x_i)}$ contains compressed information about the original $x_i$, we aim to modify this compressed $T^{(x_i)}$ to mitigate the underlying spurious correlations. The procedure is detailed as follows:

First, we employ a set of multiple classifiers $\mathcal{M}$, where each $m_i \in \mathcal{M}$ represents an individual classifier trained on $\mathcal{D_{\textit{\small{ori}}}}$. After training $\mathcal{M}$, we measure word importance on $x_i$ using each $m_i$ and extract $K$ important words, denoted as $W^{(m_i)}$. We then count the frequency of each word's appearance in the $W^{(m_i)}$. If a certain word appears in $W^{(m_i)}$ more than the threshold $\tau$\footnote{We simply set $\tau$ as $(|\mathcal{M}|+1)/2$, where $|\mathcal{M}|$ denotes the number of classifier models. Note that $|\mathcal{M}|$ is an odd number.}, indicating its importance across various models, we include it in $W_p$, the set of principal words crucial for determining the label of $x_i$. Words in $W^{(m_i)}$ that are not included in $W_p$ are categorized into $W_s$, as they are important only for certain models and not universally significant, implying that such words may induce spurious correlations in the model. Additionally, arbitrary words designated by the engineer that are known to induce spurious correlations and biases can also be included in $W_s$ in case of need.

Second, we modify $T^{(x_i)}$ to mitigate spurious correlations at the triple-level. Specifically, we first categorize each $t^{(x_i)}_j$ in $T^{(x_i)}$ as a spurious triple if $t^{(x_i)}_j$ contains a word from $W_s$. Other triples that contain a word from $W_p$ are categorized as principal triples. After categorization, we obtain $\hat{T}^{(x_i)}$ by modifying only the principal triples while maintaining the spurious triples. In particular, we use $\mathcal{L}$ to alter the label of $x_i$ by modifying the principal triples, which play a crucial role in determining the label. This process results in the generation of modified principal triple, $\hat{t}^{(x_i)}_j = \mathcal{L}(t^{(x_i)}_j, \hat{y}_i, p_{\textit{\small{mod}}})$, where $\hat{y}_i \neq y_i$ denotes the desired label. This targeted manipulation preserves the spurious words and semantics of the original data while converting the label, thereby augmenting counterbias data.

Finally, to effectively leverage the flexibility of counterbias data, which allows for various changes compared to the original data, such as different syntactic structures, we randomly permute the order of normal triples that are not categorized as spurious or principal triples within $\hat{T}^{(x_i)}$. Additionally, gender bias-inducing words are replaced with words of the different genders at the triple-level to mitigate gender bias. We used the WinoBias dataset~\cite{zhao2018gender} to perform this modification. Furthermore, we randomly delete some of the normal triples with a small, predefined probability. Rearrangement and deletion of normal triples help introduce diverse patterns into the augmented data while preserving the core semantics. The completion of this process produces the final candidate set of triples for reconstruction, $\hat{T}^{(x_i)}$.

\subsubsection{Reconstruction of Counterbias Text}

Finally, we augment counterbias text $\hat{x}_i$ by reconstructing text given the processed $\hat{T}^{(x_i)}$. Specifically, we utilize $\mathcal{L}$ to achieve this, which is formulated as $\hat{x}_i = \mathcal{L}(\hat{T}^{(x_i)}, p_{\textit{\small{rec}}})$. As a consequence, we obtain the counterbias data $(\hat{x}_i, \hat{y}_i)$. Note that we can easily generate multiple $\hat{x}_i$ using different configurations of decoding strategies for $\mathcal{L}$ or even different arrangements of $\hat{T}^{(x_i)}$. This is different from conventional counterfactual data augmentation, which is difficult to augment multiple data as they require minimal changes to the original data.

\section{Experiments}

\begin{table*}[]
\centering
\begin{adjustbox}{max width=\textwidth}
\begin{tabular}{lcccc}
\Xhline{3\arrayrulewidth}
 & \textbf{BERT} & \textbf{DeBERTaV3} & \textbf{T5} & \textbf{ModernBERT} \\ 
\hline\hline
Baseline w/o Augmentation
& 92.8 | 91.5 | 86.2 | 82.4 
& 94.0 | 91.6 | 86.6 | 84.5 
& 94.5 | 92.3 | 85.4 | 83.8 
& 93.8 | 94.1 | 88.3 | 89.8 \\

EDA \cite{wei2019eda}        
& 93.1 | 90.8 | 86.8 | 80.6 
& 93.1 | 91.6 | 86.2 | 81.5 
& 92.9 | 91.6 | 88.8 | 82.6 
& 92.4 | 91.4 | 89.7 | 86.9 \\

BT \cite{sennrich2016improving}  
& 93.2 | 91.4 | 87.7 | 83.1 
& 93.5 | 92.2 | 84.1 | 82.8 
& 89.3 | 88.2 | 88.1 | 83.4 
& 94.8 | 93.9 | 89.7 | 85.7 \\

C-BERT \cite{wu2019conditional} 
& 91.9 | 92.1 | 84.4 | 82.1 
& 94.0 | 91.0 | 89.0 | 84.7 
& 93.2 | 90.8 | 91.2 | 85.4 
& 90.1 | 90.3 | 91.1 | 89.3 \\

Human-CAD \cite{kaushik2020learning}
& \sout{00.0} | 93.2 | 88.0 | \sout{00.0} 
& \sout{00.0} | 93.8 | \underline{89.9} | \sout{00.0} 
& \sout{00.0} | 95.1 | 89.9 | \sout{00.0} 
& \sout{00.0} | 97.2 | 91.2 | \sout{00.0} \\

AutoCAD \cite{wen2022autocad} 
& \textbf{94.9} | 92.8 | 88.0 | 89.8 
& 96.4 | 93.3 | \textbf{90.1} | \underline{91.3} 
& 95.2 | 93.4 | 89.1 | \textbf{92.0} 
& 95.4 | 95.7 | 91.6 | 92.4 \\

GPT3Mix \cite{yoo2021gpt3mix}
& 93.2 | 93.9 | \sout{00.0} | \sout{00.0} 
& 95.2 | \textbf{94.1} | \sout{00.0} | \sout{00.0} 
& 95.3 | 93.9 | \sout{00.0} | \sout{00.0} 
& 96.7 | 96.3 | \sout{00.0} | \sout{00.0} \\

AugGPT \cite{dai2023auggpt} 
& 94.2 | 92.2 | \textbf{90.3} | 88.7 
& 95.4 | 94.0 | 87.5 | 87.6 
& 95.7 | \underline{94.2} | 88.9 | 85.1 
& 95.1 | 95.1 | \underline{92.1} | 89.6 \\

\textsc{CoBA} (\textit{LLM-Identification}) 
& \underline{94.6} | \underline{94.4} | 89.9 | \underline{90.6} 
& \textbf{96.7} | \underline{94.0} | 88.2 | 90.6 
& \underline{95.9} | 93.8 | \textbf{90.5} | 89.2 
& 96.3 | 96.9 | \textbf{92.2} | \textbf{93.1} \\

\textsc{CoBA} 
& \textbf{94.9} | \textbf{95.4} | \underline{90.1} | \textbf{91.1} 
& \underline{96.5} | \textbf{94.1} | \textbf{90.1} | \textbf{91.7} 
& \textbf{96.2} | \textbf{95.3} | \textbf{90.5} | \underline{91.3} 
& \textbf{96.2} | \textbf{96.9} | 91.0 | \underline{92.9} \\
\Xhline{3\arrayrulewidth}
\end{tabular}
\end{adjustbox}
\caption{Comparison of downstream task performance under different data augmentation strategies. The best performance in each group is in \textbf{bold}, and the second-best is in \underline{underline}. Performances are reported as ``SST-2 | IMDB | SNLI | MNLI'' under each model. Note that Human-CAD only provides counterfactual datasets for IMDB and SNLI, and the official source code of GPT3Mix is limited to processing large datasets such as SNLI and MNLI, which are denoted as ``\sout{00.0}''.}
\label{tab:exp-performance}
\end{table*}

\subsection{Improvement on Task Performance}

We first evaluated performance improvements in downstream tasks to determine if \textsc{CoBA} effectively mitigates spurious correlations and outperforms conventional data augmentation methods, including counterfactual data augmentation. For this purpose, we primarily used natural language inference (NLI) and sentiment analysis tasks. We used SNLI \cite{bowman2015large} and MNLI \cite{williams2018broad} for NLI tasks and SST-2 \cite{socher2013recursive} and IMDB \cite{maas2011learning} for sentiment analysis tasks, considering the configuration of previous studies \cite{wen2022autocad, wang2022identifying, treviso2023crest, li2024prompting}. Implementation details and baseline methods are provided in Appendix~\ref{app:exp-setup-task-performance}.

Table~\ref{tab:exp-performance} demonstrates the result of the experiment. A key finding is that \textsc{CoBA} outperformed other baselines, including counterfactual data augmentation methods, in most cases. Although counterfactual data augmentation methods effectively mitigate spurious patterns, they limit data diversity by introducing minimal modifications when converting labels. On the other hand, conventional data augmentation methods, particularly LLM-based methods such as GPT3Mix and AugGPT exhibit the variation in augmented data; however, they do not take spurious correlations into account. By combining the advantages of mitigating spurious correlations and generating diverse augmented data, \textsc{CoBA} was able to outperform other baselines.

Additionally, we conducted a small ablation study, where LLMs identified $W_s$ and $W_p$ instead of using $\mathcal{M}$, a set of multiple well-trained classifiers. The results are presented as ``\textsc{CoBA} (LLM-Identification)’’ in Table~\ref{tab:exp-performance}. While this approach showed remarkable performance compared to other baselines, its improvement was smaller than that of the original \textsc{CoBA}. This suggests that identifying spurious and principal words using LLM may be less effective than our majority-voting-based ensemble method with downstream task models. We hypothesize this phenomenon arises because a single LLM may not effectively capture $W_s$ and $W_p$, given the difference in important words across models, as highlighted in Section~\ref{sec:method-analysis-important}.

\subsection{Mitigation of Spurious Correlation}

\begin{table}[t]
\centering
\resizebox{0.7\columnwidth}{!}{%
\begin{tabular}{c|cc}
\Xhline{3\arrayrulewidth}
                 & IMDB          & SNLI          \\ \hline\hline
w/o Augmentation & 52.3          & 70.2          \\
AutoCAD          & 86.1          & 75.6          \\
\textsc{CoBA}    & \textbf{87.2} & \textbf{75.8} \\ \Xhline{3\arrayrulewidth}
\end{tabular}
}
\caption{The comparison of models on Human-CAD test set. For this experiment, we trained \texttt{BERT-base}.}
\label{tab:exp-spurious}
\end{table}

To verify that the effectiveness of the proposed method comes from mitigating spurious correlations rather than just data augmentation, we adopted the Human-CAD test set, which provides human-annotated examples for assessing spurious correlation mitigation. For this evaluation, we trained \texttt{BERT-base} using a combination of $\mathcal{D}_{\textit{\small{ori}}}$ and $\mathcal{D}_{\textit{\small{cb}}}$ augmented by \textsc{CoBA} and AutoCAD.

As shown in Table~\ref{tab:exp-spurious}, the model trained without $\mathcal{D}_{\textit{\small{cb}}}$ exhibited significantly lower performance on the Human-CAD test set, indicating that the baseline model is vulnerable to spurious correlations. In contrast, models trained with AutoCAD and \textsc{CoBA} demonstrated more robust performance than the baseline. Furthermore, \textsc{CoBA}, our proposed method based on counterbias augmentation, outperformed existing counterfactual data augmentation methods, underscoring its effectiveness in mitigating spurious correlations.

\subsection{Alleviation of Gender Bias}
\label{sec:experiment-alleviation-gender-bias}

To verify \textsc{CoBA}’s effectiveness in reducing biases by mitigating related spurious correlations, we conducted an experiment focused on gender bias reduction. For this experiment, we adopted the list of gender bias-related words from a previous study \cite{zhao2018gender}. By incorporating words from this list into $W_s$, we aim to mitigate the underlying spurious correlations, thereby alleviating gender bias in the model. To quantify the gender bias, we used three benchmarks: StereoSet (SS) \cite{nadeem2021stereoset}, CrowS-Pairs \cite{nangia2020crows}, and WikiBias \cite{zhong2021wikibias}. For comparison, we established various baselines. BERT (Raw) refers to the original BERT model without any additional training, and BERT (IMDB) refers to the BERT model with additional pretraining on the IMDB training dataset. SentenceDebias is a baseline method that achieves debiasing at the embedding level \cite{liang2020towards}. Naive-masking and Random-phrase-masking are methods based on word-level substitution \cite{thakur2023language}. Lastly, \textsc{CoBA}-PT refers to the BERT model with additional pretraining on the IMDB training dataset, combined with the augmented data generated by \textsc{CoBA}.

Table~\ref{tab:exp-gender} shows the result of the experiment. The model trained with a combination of original and CoBA-augmented counterbias data achieved a score closest to the ideal 50 on SS and CrowS-Pairs, as well as highest performance on WikiBias, outperforming other baselines. Unlike other strategies, such as masking gender-related pronouns to neutral pronouns, our \textsc{CoBA} focuses on augmenting data with representations of the different gender, leading to a more balanced introduction of gender-related representations. As a result, \textsc{CoBA} contributed to mitigating spurious correlations, thereby alleviating gender bias in the model.

\begin{table}[t]
\centering
\resizebox{0.9\columnwidth}{!}{%
\begin{tabular}{c|cc|c}
\Xhline{3\arrayrulewidth}
                                                                                          & SS            & CrowS            & WikiBias\\ \hline\hline
BERT (Raw)                                                                                & 57.8          & 59.0             & 69.0 \\
BERT (IMDB)                                                                               & 58.6          & 59.7             & 69.1 \\
\begin{tabular}[c]{@{}c@{}}SentenceDebias\\ \cite{liang2020towards}\end{tabular}          & 53.8          & 58.1             & 72.1 \\
\begin{tabular}[c]{@{}c@{}}Naive-masking\\ \cite{thakur2023language}\end{tabular}         & 56.5          & 60.8             & 71.8 \\
\begin{tabular}[c]{@{}c@{}}Random-phrase-masking\\ \cite{thakur2023language}\end{tabular} & 54.5          & 58.0             & 71.2 \\
\begin{tabular}[c]{@{}c@{}}\textsc{CoBA}-PT \end{tabular}                                 & \textbf{51.4} & \textbf{52.0}    & \textbf{75.4} \\ \Xhline{3\arrayrulewidth}
\end{tabular}
}
\caption{Comparison of gender bias across methods, as measured by SS and CrowS. A score close to 50 for SS and CrowS, along with a higher score for WikiBias, indicates less gender bias in the model. Note that all models are based on \texttt{BERT-base}.}
\label{tab:exp-gender}
\end{table}

\subsection{OOD Robustness with Diverse Augmented Data}

\begin{table}[t]
\centering
\resizebox{0.75\columnwidth}{!}{%
\begin{tabular}{c|c|c|c}
\Xhline{3\arrayrulewidth} 
EDA    & AutoCAD & AugGPT &\textsc{CoBA} \\ \hline \hline
0.9957 & 0.9641  & 0.9658 & 0.9531  \\ \Xhline{3\arrayrulewidth}
\end{tabular}%
}
\caption{Cosine similarity between the embedding vectors of $x_i$ and $\hat{x}_i$ from each method.}
\label{tab:exp-sample-embedding}
\end{table}

Unlike counterfactual text augmentation, which introduces minimal modifications to alter labels, counterbias text augmentation has no such restrictions, allowing for a wider range of lexical and semantic expressions. This flexibility plays a crucial role in enhancing model performance through data augmentation \cite{cegin2024effects}. To validate this, we randomly sampled 100 augmented data from IMDB using each method and measured the difference between the original and augmented data by calculating the cosine similarity produced by \texttt{BERT-base}. The results in Table~\ref{tab:exp-sample-embedding} indicate that \textsc{CoBA} introduces meaningful differences in the data while preserving core semantics.

To further support the effectiveness of this diversification in augmented data, we conducted an evaluation in an OOD scenario. For this experiment, we trained a model on IMDB but tested it on SST-2 and Yelp \cite{zhang2015character}. The results of this evaluation are presented in Table~\ref{tab:exp-ood}. This evaluation suggests that our \textsc{CoBA} exhibits remarkable improvement in OOD robustness. While the performance gain is slightly lower than that of ReAct \cite{sun2021react} baseline, it is important to note that ReAct is a strategy that solely focused on enhancing OOD robustness. In contrast, our \textsc{CoBA} offers various benefits such as mitigation of spurious correlation and other biases. In conclusion, we validated that \textsc{CoBA} jointly offers numerous benefits to the model, from the mitigation of spurious correlation to the improvement of OOD robustness.

\begin{table}[t]
\centering
\resizebox{0.9\columnwidth}{!}{%
\begin{tabular}{c|cc}
\Xhline{3\arrayrulewidth}
                                                                    & IMDB → SST-2 & IMDB → Yelp\\ \hline\hline
Baseline                                                            & 63.2        & 61.2        \\
EDA                                                            & 66.2       & 58.2        \\
AutoCAD                                                            & 80.1        & 73.6        \\
AugGPT                                                            & 83.0        & 71.6        \\
ReAct                                                             & 84.5        & 75.3        \\
\textsc{CoBA}                                                    & 83.2        & 74.0        \\ \Xhline{3\arrayrulewidth}
\end{tabular}
}
\caption{AUROC(\%) performance of the models in OOD scenario. ``IMDB → SST-2'' indicates a scenario where a model trained on IMDB is tested on SST-2, and ``IMDB → Yelp'' means the model trained on IMDB is tested on Yelp. For ReAct, we follow the reported performance from previous work \cite{baran2023classical}.}
\label{tab:exp-ood}
\end{table}

\subsection{Extension to Generation Tasks}
\begin{table}[t]
\centering
\resizebox{0.85\columnwidth}{!}{%
\begin{tabular}{c|c|c}
\Xhline{3\arrayrulewidth}  
Baseline & AugGPT & \textsc{CoBA}  \\ \hline \hline
72.10 \& 73.59  & 74.50 \& 75.21 & 75.70 \& 76.08  \\ \Xhline{3\arrayrulewidth} 
\end{tabular}%
}
\caption{BLEU-4 scores \cite {papineni2002bleu} for the informal-to-formal text style transfer task on the GYAFC dataset. We used for \texttt{T5-Base} \cite{raffel2020exploring} and \texttt{Flan-T5-Base} \cite{chung2024scaling} for this experiment.}
\label{tab:exp-tst}
\end{table}

In this paper, we proposed \textsc{CoBA}, which involves decomposing the given text into semantic triples, selecting spurious and principal triples, applying bias-mitigation techniques, and then reconstructing the augmented text. This approach is applicable not only to classification tasks but also to text generation tasks. To verify \textsc{CoBA}'s effectiveness in text generation, we conducted an experiment applying \textsc{CoBA} to a text style transfer task using the GYAFC dataset~\cite{rao2018dear}.

For applying \textsc{CoBA} to the text style transfer task, we first decomposed the given $x_i$ into $T^{(x_i)}$. Next, we utilized $\mathcal{L}$ to identify principal triples in both formal and informal sentences\footnote{Note that our purpose in this experiment is to verify the usefulness of \textsc{CoBA}, a triple-based augmentation method, in text generation tasks, rather than to mitigate spurious correlations in these tasks. Accordingly, we did not identify spurious triples. We leave the identification and mitigation of spurious correlations in text generation tasks as future work.}. Additionally, we included the gender bias alleviation scheme introduced in Section~\ref{sec:experiment-alleviation-gender-bias}. After permuting the order of normal triples that are not principal, we reconstructed the augmented text.

The results of this experiment, displayed in Table~\ref{tab:exp-tst}, show that \textsc{CoBA} exhibited a remarkable performance improvement compared to the AugGPT baseline. This underscores \textsc{CoBA}’s extensibility to text generation tasks based on the triple-level modifications. We plan to investigate the strategies for identifying spurious patterns in text generation tasks in future work.

\begin{table}[]
\centering
\resizebox{\columnwidth}{!}{%
\begin{tabular}{c|c}
\Xhline{3\arrayrulewidth} 
           & Original [Label: Neutral]                                                                                                                                               \\ \hline \hline
Premise    & \begin{tabular}[c]{@{}c@{}}A \textbf{woman} talks on a cellphone \\ while sitting in front of \\ blue railings that are in front of the ocean.\end{tabular}                      \\ \hline
Hypothesis & \textbf{She} talks to her boyfriend about plans that night.                                                                                                                      \\ \Xhline{2\arrayrulewidth} 
           & Human-CAD [Label: Contradiction]                                                                                                                                        \\ \hline \hline
Premise    & \begin{tabular}[c]{@{}c@{}}A \textbf{woman} talks on a cellphone \\ while sitting in front of \\ blue railings that are in front of the ocean.\end{tabular}                      \\ \hline
Hypothesis & \textbf{He} has a conversation on her phone outdoors.                                                                                                                            \\ \Xhline{2\arrayrulewidth} 
           & \textsc{CoBA} [Label: Contradiction]                                                                                                                                    \\ \hline \hline
Premise    & \begin{tabular}[c]{@{}c@{}}A \textbf{man} is sitting in front of \\ blue railing while talking on a cellphone,\\ with the railing positioned in front of the ocean.\end{tabular} \\ \hline
Hypothesis & \textbf{A man} talks to his boyfriend while he is in a new car.                                                                                                                  \\ \Xhline{3\arrayrulewidth} 
\end{tabular}%
}
\caption{The comparison of augmented data generated by Human-CAD and \textsc{CoBA} on SNLI.}
\label{tab:qualitative}
\end{table}

\subsection{Qualitative Analysis}
\label{sec:experiment-qualitative}

Table~\ref{tab:qualitative} compares Human-CAD and CoBA for data augmentation in the SNLI dataset. In this example, the relationship between the original premise and hypothesis is neutral, while the augmented pairs exhibit a contradiction. In the Human-CAD example, the premise remains unchanged, and the label change is introduced by altering the gender in the hypothesis. Conversely, \textsc{CoBA} modifies “outdoor” to “car” to change the label to a contradiction, leading to a contextually meaningful change without human annotation. Additionally, \textsc{CoBA}’s modification of the premise enhances the diversity of augmented data, likely contributing to the performance improvements shown in Table~\ref{tab:exp-performance}. Additional qualitative analysis results can be found in Appendix \ref{app:exp-qualitative-coba}.


\section{Conclusion}

We introduced \textit{counterbias data augmentation} as a more general and flexible extension of counterfactual data augmentation, capable of addressing multiple forms of bias and improving out-of-distribution robustness. Through an analysis of word importance across different models, we highlighted the limitations of using a single model to identify spurious correlations. Building on these insights, we developed \textsc{CoBA}, a framework that leverages LLMs to decompose text into semantic triples and apply triple-level modifications guided by a majority-voting-based ensemble. This approach enabled us to effectively mitigate spurious correlations and alleviate biases.

Our extensive experiments demonstrated \textsc{CoBA}’s versatility and effectiveness across various tasks, including sentiment analysis, natural language inference, and text style transfer. Unlike existing counterfactual methods that emphasize minimal label-flipping modifications, \textsc{CoBA} allows for more diverse and semantically rich augmentations, leading to broader improvements in both accuracy and robustness.

Another potential strength of \textsc{CoBA} is its explainability through semantic triples. Since \textsc{CoBA} generates triples as an intermediate step during augmentation, it provides a clearer view of LLM behavior in text augmentation. Additionally, analyzing and comparing triples extracted from different LLMs can deepen our understanding of their mechanisms. Future research could further explore this direction to enhance LLM explainability.

Looking ahead, we plan to extend \textsc{CoBA} to more complex text generation scenarios, further exploring the framework’s potential to mitigate spurious patterns in generated text. We envision future work refining the decomposition and reconstruction steps, optimizing the balance between information preservation and bias mitigation, and generalizing our approach to a wider range of application domains and model architectures. 

\section*{Limitation}
We discuss several limitations of our approach. While the \textsc{CoBA} framework demonstrates promising results in mitigating spurious correlations and various biases, it is important to acknowledge the inherent challenges and trade-offs associated with our methodology. In this study, we acknowledge several limitations:

\paragraph{Potential Information Loss in Semantic Triples.} Our triple-based approach to text representation, although effective in capturing core semantic relationships, inherently entails some degree of information loss. Decomposing text into subject-predicate-object triples simplifies complex linguistic structures and potentially discards contextual nuances that could be relevant for certain tasks. This simplification, while advantageous for text manipulation and reconstruction, may inadvertently introduce new patterns that manifest as alternative forms of spurious correlations.

\paragraph{Trade-off Between Semantic Preservation and Flexible Augmentation.} The intentional sacrifice of some semantic details is a design choice to foster diversity in the augmented data and mitigate spurious correlations. However, this compromise may not be ideal for all domains and tasks, especially where fine-grained information is essential. Furthermore, the robustness of our majority-voting ensemble method for identifying spurious correlations, while validated in our tested scenarios, may vary when applied to different types of biases or domains beyond the scope of our current evaluation.

\paragraph{Accuracy of Triple-level Modifications.} We acknowledge that our study did not perform an extensive evaluation regarding the accuracy of triple decomposition and reconstruction through LLMs. Instead, we focused on empirical validation of performance improvement and bias mitigation through augmented counterbias data, thereby validating the effectiveness of \textsc{CoBA}. Nonetheless, it is important to recognize that the accuracy of these triple-level modifications may vary, and any inaccuracies in decomposition or reconstruction could potentially affect the downstream task performance. Future work should include a more granular evaluation of the fidelity of the extracted semantic triples and the consistency of the reconstructed text, as well as the development of robust metrics to quantify the precision of these modifications.

\paragraph{Limited Scope of Addressed Biases.} Although our study is one of the first studies to alleviate multiple biases through counterbias data augmentation, our analysis primarily focuses on specific biases within the datasets examined. Although our framework has demonstrated success in these contexts, its generalizability to other types of biases remains to be fully explored. For instance, future studies in this direction could incorporate a validation on racial bias and concept bias \cite{field2021survey, zhou2024explore}.

\paragraph{Limited Usage of POS Tagging.} Our approach utilizes LLMs for triple decomposition. While lightweight POS taggers could offer computational efficiency, they operate at the sentence level and struggle with complex structures. We considered comparing these methods but found that lightweight approaches often failed to extract meaningful triples from long or intricate sentences. In contrast, LLMs capture higher-order semantics beyond sentence boundaries. However, reducing computational overhead remains crucial. Future work could explore hybrid approaches, leveraging lightweight methods when feasible and applying LLMs selectively for complex cases to balance efficiency and accuracy in \textsc{CoBA}.

\paragraph{Limitations of Majority-Voting Strategy.} While the majority-voting strategy employed in \textsc{CoBA} has shown strong performance in aggregating outputs from diverse models, it may overlook signals from particularly robust individual models in scenarios where model robustness is imbalanced. Although such extreme cases were not observed in our experiments, we recognize this as a potential limitation, particularly in other domains or under different configurations. Future work may explore adaptive voting mechanisms or reliability-weighted aggregation strategies to better account for model-specific performance differences.

\section*{Ethics Statement}
In this study, the replacement of gender-related words in \textsc{CoBA} was based on a binary conceptualization of gender, excluding non-binary identities and other complex gender expressions. This limitation arose because the gender-related word list we used from a previous study was restricted to binary genders \cite{zhao2018gender}. While this approach helped systematize the experiments, it inevitably overlooked the nuances of gender diversity. As a result, the framework’s binary focus presents an ethical limitation. We fully recognize the importance of addressing this issue and hope future research will expand counterbias data augmentation to inclusively represent non-binary genders.

\section*{Acknowledgement}
This work was supported by the Institute of Information \& Communications Technology Planning \& Evaluation (IITP) grant funded by the Korea government (MSIT) [RS-2021-II211341, Artificial Intelligence Graduate School Program (Chung-Ang
University)] and by the National Research Foundation of Korea (NRF) grant funded by the Korea
government (MSIT) (RS-2025-00556246).


\appendix
\section{Experimental Setup for Word-level Importance Analysis}
\label{app:exp-setup-word-importance}


For experiments on important word analysis in Section \ref{sec:method-analysis-important}, we employed four models—\texttt{BERT-base}, \texttt{BERT-large}~\cite{devlin2019bert}, \texttt{RoBERTa-large}~\cite{liu2019roberta}, and \texttt{DistilBERT}~\cite{sanh2019distilbert} to identify the top-importance words. For training the classifier, we set the batch size to 32, the initial learning rate of the AdamW optimizer~\cite{loshchilov2019decoupled} to 5e-5, the maximum token length to 300, and the maximum training epochs to 15. We selected the best checkpoint based on the accuracy of the validation set to extract the word importance of each model.

Additionally, the second analysis involves performing POS tagging using NLTK \cite{loper2002nltk} and comparing the tendencies of each model concerning the POS tags of important words. Specifically, we used pre-trained English Punkt tokenizer \cite{kiss2006unsupervised}. For this analysis, we employed four different models: \texttt{BERT-base}, \texttt{BERT-large} \cite{devlin2019bert}, \texttt{RoBERTa-large} \cite{liu2019roberta}, and \texttt{DistilBERT} \cite{sanh2019distilbert}. We trained these models on SST-2 \cite{socher2013recursive} and IMDB \cite{maas2011learning} datasets.

\section{Experimental Setup for Task Performance}
\label{app:exp-setup-task-performance}
We employed five models: \texttt{BERT-base}, \texttt{BERT-large} \cite{devlin2019bert}, \texttt{RoBERTa-large} \cite{liu2019roberta}, \texttt{DistilBERT} \cite{sanh2019distilbert}, and \texttt{BART-base} \cite{lewis2020bart} to identify $w_s$ and $w_p$. In this experiment, we set the number of top-k words to 5. We used \texttt{GPT-4o-mini} \cite{openai2024gpt4omini} for triple decomposition and text reconstruction. The generated counterfactual data was combined with the original dataset and used to train \texttt{BERT-base}, \texttt{DeBERTaV3-base}~\cite{he2023debertav3}, \texttt{T5-base}~\cite{raffel2020exploring} and \texttt{ModernBERT}~\cite{warner2024smarter} classifiers. For training the classifiers, we set the batch size to 32, the initial learning rate of the AdamW optimizer~\cite{loshchilov2019decoupled} to 5e-5, and the maximum training epochs to 10. The best checkpoint was selected based on validation set accuracy. All experiments were conducted using the Transformers library \cite{wolf2020transformers}.

For evaluating task performance, the experiments in this paper utilize datasets from two domains: sentiment analysis and SNLI. Specifically, for sentiment analysis, the SST-2 (Stanford Sentiment Treebank) and IMDB datasets are employed, both of which belong to the movie review domain and feature binary classification labels, positive or negative sentiment. The IMDB dataset \cite{maas2011learning}, consists of approximately 50,000 reviews, while SST-2 \cite{socher2013recursive} contains around 67,000 sentences. For NLI tasks, the SNLI \cite{bowman2015large} and MNLI \cite{williams2018broad} datasets are used. These datasets include three classification labels: contradiction, neutral, and entailment. SNLI comprises around 570,000 sentence pairs derived from image captions. MNLI  contains approximately 433,000 sentence pairs spanning multiple domains, making it a more diverse and challenging benchmark. The experiments leverage the Datasets library \cite{lhoest2021datasets} to access and preprocess these datasets, ensuring consistency and ease of implementation across different models.

The baseline methods for comparison are as follows:

\begin{itemize}
  \item EDA \cite{wei2019eda}: A rule-based augmentation technique that modifies sentences through word-level modification. In this study, the modification ratio was set to 20\%.
  \item Back-translation \cite{sennrich2016improving}: An augmentation technique that translates the original sentence into a pivot language and then back-translates it into the source language.
  \item C-BERT \cite{wu2019conditional}: A strategy that leverages the contextual capabilities of the BERT model by filling in masked tokens.
  \item Human-CAD \cite{kaushik2020learning}: This baseline uses the Human-CAD dataset, which was created by employing human annotators to generate counterfactual data from a subset of the SNLI and IMDB datasets. Specifically, we trained a model using a combination of the Human-CAD dataset and the original dataset.
  \item AutoCAD \cite{wen2022autocad}: A counterfactual data augmentation method that uses a text-infilling model.
  \item GPT3Mix \cite{yoo2021gpt3mix}: An LLM-based augmentation technique using few-shot examples and the assignment of soft label predicted by the LLM. We used \texttt{GPT-4o-mini} for a fair comparison.
  \item AugGPT \cite{dai2023auggpt}: An augmentation approach based on ChatGPT, where LLMs are prompted to generate paraphrases of original sentences. We used \texttt{GPT-4o-mini} for a fair comparison.
\end{itemize}

\section{Ablation Study on \textsc{CoBA}}
\label{app:exp-ablation}

\begin{table}[]
\centering
\resizebox{0.7\columnwidth}{!}{%
\begin{tabular}{c|c|c}
\hline
\Xhline{3\arrayrulewidth}
\begin{tabular}[c]{@{}c@{}}Gender Bias\\ Word Change\end{tabular} & \begin{tabular}[c]{@{}c@{}}Triple\\ Shuffling\end{tabular} &      \\ \hline \hline
                                                                  &                                                            & 94.9 \\ \hline
                                                                  & \ding{51}                                                          & 95.7 \\ \hline
\ding{51}                                                                 &                                                            & 92.9 \\ \hline
\ding{51}                                                                 & \ding{51}                                                          & 95.4 \\ \hline
\Xhline{3\arrayrulewidth}
\end{tabular}%
}
\caption{Ablation study results for \textsc{CoBA}. `\ding{51}' indicates that the specified method was applied, while blank indicates that the specified method was not performed. The baseline model at the top refers to the method where the obtained triples are directly reconstructed without any modifications.}
\label{tab:app-ablation}
\end{table}

\begin{table}[t]
\centering
\resizebox{\columnwidth}{!}{%
\begin{tabular}{lcc}
\toprule
\# of Models & SST-2 Accuracy & IMDb Accuracy \\
\midrule
1 & 91.8 & 93.5 \\
3 & 91.8 & 93.1 \\
5 & \textbf{94.9} & \textbf{95.4} \\
7 & 94.0 & 94.1 \\
\bottomrule
\end{tabular}
}
\caption{Performance based on the number of models used in the ensemble.}
\label{tab:ensemble-size}
\end{table}

\begin{table}[t]
\centering
\resizebox{\columnwidth}{!}{%
\begin{tabular}{lcc}
\toprule
Top-\textit{k} Words & SST-2 Accuracy & IMDb Accuracy \\
\midrule
1 & 90.7 & 92.2 \\
3 & 91.8 & 91.7 \\
5 & \textbf{94.9} & \textbf{95.4} \\
10 & 94.1 & 94.9 \\
20 & 93.5 & 92.8 \\
\bottomrule
\end{tabular}
}
\caption{Performance based on number of top-\textit{k} words selected per model.}
\label{tab:topk-words}
\end{table}

\begin{table*}[t!]
\centering
\resizebox{.8\textwidth}{!}{%
\begin{tabular}{c}
\hline 
\Xhline{3\arrayrulewidth}
Original [Label: Negative]                                                                                                                                                                                                                                                                                                                                                                                                                                                                                                                                                                                                                                    \\ \hline
\begin{tabular}[c]{@{}c@{}}Penn takes the time to develop his characters, and we almost care about them.\\  However there are some real problems with the story here, \\ we see no real motivation for the evil brother's behavior, and the time line is screwed up. \\ Supposedly set in 1963, the music is late 60s/early 70s. \\ The references and dialogue is 70s/80s. \\ The potential for a powerful climax presents itself, and Penn allows it to slip away. \\ But even with all these difficulties it is worth the watch, but not great.\end{tabular}                                                                              \\ \hline
Human-CAD [Label: Positive]                                                                                                                                                                                                                                                                                                                                                                                                                                                                                                                                                                                                                                   \\ \hline
\begin{tabular}[c]{@{}c@{}}Penn takes the time to develop his characters, and we truly care about them. \\ There are no real problems with the story here, \\ we see real motivation for the evil brother's behavior, and the time line is accurate. \\ Set in 1963, the music is early 60s. The references and dialogue are 60s too. \\ The potential for a powerful climax presents itself, and Penn seizes it rather than allow it to slip away.\\  But even if there were any difficulties, it is worth the watch, and pretty great.\end{tabular}                                                                                     \\ \hline
\textsc{CoBA} [Label: Positive]                                                                                                                                                                                                                                                                                                                                                                                                                                                                                                                                                                                                                                        \\ \hline
\begin{tabular}[c]{@{}c@{}}The timeline is screwed up, and the references and dialogue are 70s/80s, \\ even though the story is supposedly set in 1963. \\ Despite these issues, there are some interesting elements in the story. \\ Penn takes time to develop his characters, \\ and the potential for a powerful climax presents itself, which Penn delivers effectively. \\ Overall, it is great, although the music is from the late 60s/early 70s. \\ It is definitely worth the watch, even though we see no motivation \\ for the evil brother's behavior. \\ Nevertheless, we genuinely care about the characters.\end{tabular} \\ \hline \Xhline{3\arrayrulewidth}
\end{tabular}%
}
\caption{The comparison of augmented data generated by Human-CAD and \textsc{CoBA} on IMDB.}
\label{tab:app-qualitative1}
\end{table*}

To verify the effectiveness of the proposed method, we conducted an ablation study using \texttt{BERT-base} trained with \textsc{CoBA} on IMDB dataset. For this experiment, we systematically removed one or more components of our proposed approach to assess the contribution of each component. 

The results of this experiment is presented in Table \ref{tab:app-ablation}. We observed that solely replacing gender-related words led to a decline in performance. However, when shuffling was introduced, performance not only improved beyond the baseline but also surpassed our proposed \textsc{CoBA}, highlighting the significant impact of augmented data diversity on model performance. Nevertheless, it is important to note that \textsc{CoBA} is designed to simultaneously enhance model performance and mitigate underlying biases, providing multiple advantages for model training. Additionally, the modular structure of \textsc{CoBA} enables engineers to selectively utilize its components. For example, if the training dataset is confirmed to be free of significant biases, engineers can exclude the replacement procedure and use only shuffling, thereby maximizing the performance gains from data augmentation.

Also, We investigated the effect of the number of models used in the ensemble on downstream performance. The goal was to determine how model diversity impacts the robustness of generated counterfactuals. As shown in Table~\ref{tab:ensemble-size}, increasing the ensemble size to five improves accuracy, while performance slightly saturates or even drops beyond that. This suggests that our choice of using five diverse models is empirically justified and not arbitrary.

In addition, we conducted an ablation study on the number of top-ranked words selected from each model when generating counterfactuals. As shown in Table~\ref{tab:topk-words}, selecting the top-5 words per model yields the highest accuracy across both datasets. Choosing too few limits diversity, while too many introduces noise, validating our design choice of top-5 selection as a balanced and effective configuration.

\section{Additional Analysis}
\label{app:exp-qualitative}

\subsection{Qualitative Analysis on \textsc{CoBA}}
\label{app:exp-qualitative-coba}

In addition to the qualitative analysis in Section~\ref{sec:experiment-qualitative}, we provide further examples. Table \ref{tab:app-qualitative1} compares augmented data generated by Human-CAD and \textsc{CoBA} using the IMDB dataset. The original data and its Human-CAD augmentations, modified by human annotators, demonstrate that while the overall syntactic structure of the text is preserved, key expressions are systematically altered. Notably, negative expressions are explicitly transformed. For instance, the phrase “time line is screwed up” was modified to “time line is accurate.” However, it should be noted that the most crucial phrase determining sentiment in this sentence is “not great.”

Whereas Human-CAD modifies most expressions that contribute to the label, \textsc{CoBA} instead preserves less critical phrases like “time line is screwed up” and focuses on altering key sentiment-related information, such as changing “Overall, it is not great” to “Overall, it is great.” This targeted modification ensures that the augmented data remains close to the local decision boundary, thereby enhancing model performance \cite{gardner2020evaluating}. Moreover, by shuffling the order of triples before reconstructing the text, \textsc{CoBA} generates more syntactically diverse expressions compared to Human-CAD.

\subsection{Qualitative Analysis on POS Tagging Analysis}
\label{app:exp-qualitative-pos}

Table \ref{tab:word-pos} presents the differences in POS tag distributions across various models. To examine how each model assigns POS tags to principal words, we provide examples of principal words selected by three models using IG \cite{sundararajan2017axiomatic} in Table \ref{tab:app-qualitative2}. A notable finding is that the \texttt{BERT-base} model identified the proper noun “Nancy” as the most important word. This aligns with prior research indicating that certain proper nouns are often associated with specific labels, leading to spurious correlations \cite{wang2020identifying}. Analyzing sentences from the IMDB dataset containing “Nancy”, we found that approximately 70

Typically, words like “enjoy” or “entertaining” are considered crucial for sentiment analysis. However, not all models prioritized these words. For example, the \texttt{RoBERTa-base} model identified “succeeded” as a principal word. While “succeeded” can imply positivity, it carries little significance in the given sentence. This variation suggests that each model exhibits different tendencies toward spurious correlations, leading to inconsistencies in POS tagging for important words and demonstrating divergent patterns across models.

\begin{table}[]
\centering
\resizebox{0.9\columnwidth}{!}{%
\begin{tabular}{c}
\Xhline{3\arrayrulewidth} 
Original \\
\hline
\begin{tabular}[c]{@{}c@{}}I really enjoyed this movie. \\ It succeeded in doing something that few movies do now; \\ it provided family values while entertaining me. \\ Nancy Drew is a heroine for all generations \\ and a role model for young girls to look up to.\end{tabular} \\ \hline
\texttt{BERT-base} \\
\hline
\begin{tabular}[c]{@{}c@{}}I really enjoyed this \textbf{movie}. \\ It succeeded in doing something that few movies do now; \\ it provided family values while \textbf{entertaining} me. \\ \textbf{Nancy} Drew is a heroine for all generations \\ and a role model for young girls to look up to.\end{tabular} \\ \hline
\texttt{DistilBERT} \\
\hline
\begin{tabular}[c]{@{}c@{}}I really \textbf{enjoyed} this movie. \\ It succeeded in doing something that few movies do now; \\ it provided \textbf{family} values while \textbf{entertaining} me. \\ Nancy Drew is a heroine for all generations \\ and a role model for young girls to look up to.\end{tabular} \\ \hline
\texttt{RoBERTa-base} \\
\hline
\begin{tabular}[c]{@{}c@{}}I really \textbf{enjoyed} this movie. \\ It \textbf{succeeded} in doing something that few movies do now; \\ it provided \textbf{family} values while entertaining me. \\ Nancy Drew is a heroine for all generations \\ and a role model for young girls to look up to.\end{tabular} \\ \hline
\Xhline{3\arrayrulewidth}
\end{tabular}%
}
\vspace{0.2cm}
\caption{Principal words for each models, \texttt{BERT-base}, \texttt{DistilBERT}, and \texttt{RoBERTa-base}.}
\label{tab:app-qualitative2}
\end{table}

\subsection{Discussion Regarding POS Tagging Analysis}

In our POS tagging analysis in Section \ref{sec:method-analysis-important-pos}, we acknowledge that dataset characteristics significantly impact the observed distributions of POS tags. Specifically, we found that the IMDB dataset exhibited a notably high frequency of the “Others” category, which can be attributed to the nature of the dataset itself. Unlike structured datasets such as SST-2, IMDB consists of lengthy internet comments, which often contain abbreviations, informal language, and exaggerated expressions. These linguistic traits contribute to a greater proportion of words that do not fall neatly into standard noun, verb, adjective, or adverb categories, thereby inflating the “Others” category.

However, despite this dataset-induced variation, we also observed model-specific differences in POS tagging tendencies. For instance, while all models predominantly identified nouns as the most frequent POS category in the SST-2 dataset, RoBERTa exhibited a relatively higher proportion of nouns in IMDB compared to other models, whereas BART demonstrated a more evenly distributed POS ratio across categories. This suggests that, although dataset properties play a crucial role in shaping POS distributions, model architectures also influence how certain word types are emphasized.

To further clarify the composition of the “Others” category, we conducted an in-depth analysis of the IMDB dataset. We found that determiners, numerals, and WH-words (e.g., “who,” “what,” “where”) appeared in similar proportions within this category. Since no single POS type overwhelmingly dominated among these, we originally grouped them under “Others” to maintain clarity. In the revised version of this work, we explicitly define which POS types are included in this category to enhance interpretability.

\section{Preliminary Evaluation on Racial Bias.}  
To assess CoBA's effectiveness beyond gender and spurious correlation biases, we conducted a preliminary experiment on the Measuring Hate Speech Corpus, which includes racially biased language. We evaluated the downstream classification accuracy of a DeBERTaV3 model under various augmentation strategies. As shown in Table~\ref{tab:racial-bias}, CoBA outperformed both traditional and LLM-based baselines, indicating its potential applicability to racial bias mitigation as well.

\begin{table}[t]
\centering
\begin{tabular}{lc}
\toprule
\textbf{Method} & \textbf{Accuracy (\%)} \\
\midrule
Baseline w/o Augmentation & 91.1 \\
EDA                       & 90.7 \\
AugGPT                    & 92.2 \\
\textbf{CoBA}             & \textbf{92.6} \\
\bottomrule
\end{tabular}
\caption{Performance on Measuring Hate Speech Corpus using DeBERTaV3.}
\label{tab:racial-bias}
\end{table}

\section{Diversity Analysis Using PCA.}  
To further evaluate how triple permutation affects the diversity of augmented data, we performed a principal component analysis (PCA) on the embeddings of generated samples from each augmentation method. We used OpenAI's \texttt{text-embedding-3-small} model to obtain sentence-level representations and measured the proportion of variance explained by the top 50 principal components. A lower explained variance suggests higher dispersion in the embedding space and thus greater data diversity. As shown in Table~\ref{tab:pca-diversity2}, CoBA achieved the most diverse outputs among the compared methods, including both traditional and LLM-based baselines.

\begin{table}[t]
\centering
\begin{tabular}{lcc}
\toprule
\textbf{Method} & \textbf{Variance (\%)} \\
\midrule
EDA        & 73.6 \\
C-BERT     & 72.1 \\
GPT3Mix    & 65.1 \\
AugGPT     & 64.9 \\
\textbf{CoBA}       & \textbf{64.3} \\
\bottomrule
\end{tabular}
\caption{Proportion of variance explained by the top 50 principal components (lower is better).}
\label{tab:pca-diversity2}
\end{table}

\section{Cost Analysis}

In our experiments, we used GPT-4o-mini, which is priced at 0.15 USD per million input tokens and 0.60 USD per million output tokens. While our approach involves two stages—semantic triple decomposition and sentence reconstruction—we conducted a detailed cost analysis using the actual prompts and the tiktoken tokenizer~\footnote{\url{https://github.com/openai/tiktoken}}. On average, the decomposition step consumes approximately 250 tokens, resulting in a cost of roughly 0.0005 USD per sample.

When applied to a dataset like IMDB, the total cost adds up to only around 2 USD, which is comparable to other LLM-based augmentation methods such as AugGPT and GPT3Mix. Although we acknowledge that our method does incur a slightly higher cost (about 2 USD more), we believe that this is a reasonable trade-off given the stronger bias mitigation and robustness improvements demonstrated in our results. 

\newpage
\section{Prompt Design}
\label{app:prompt-design}
\subsection{Prompt for Semantic Triple Decomposition (Sentiment Analysis)}

\begin{minipage}[t]
{\linewidth}\raggedright
\setlength{\parindent}{0cm}
\hrule
\vspace{1mm}

\small{
\texttt{\textcolor{blue}{System:}} You are a chatbot used for data augmentation. I will provide two paragraphs or internet comments for natural language understanding (NLU) tasks or sentiment analysis tasks. 

\texttt{\textcolor{blue}{User:}} Please create semantic triples for the following sentence. \\ Triple consists of three elements: subject, predicate, and object. \newline 

Here is an example of a sentence and its corresponding Semantic Triplet:  A few people in a restaurant setting, one of them is drinking orange juice. \newline

1. A few people | are in | a restaurant setting \\
2. One person | is drinking | orange juice \newline

Here is another example of a sentence and its corresponding Semantic Triplet:  A poor work that failed to provide a proper narrative for the black woman. \\

1. A work | is | poor \\
2. A work | failed to provide | a proper narrative \\
3. A proper narrative | is for | the black woman \newline

Please provide no answers other than the semantic triplet. Output only the semantic triplet. \newline

Here is a paragraph you should make a semantic triplet: \newline
\textcolor{gray}{\# Content}
}
\vspace{1mm}
\hrule
\end{minipage}

\vfill

\subsection{Prompt for Reconstructing Triples into Sentences (Sentiment Analysis)}

\begin{minipage}[t]
{\linewidth}\raggedright
\setlength{\parindent}{0cm}
\hrule
\vspace{1mm}

\small{
\texttt{\textcolor{blue}{System:}} You are a chatbot used for data augmentation. Your job is reconstructing the selected triples into a sentence or paragraph.

\texttt{\textcolor{blue}{User:}} Please create sentences for the following Triples. \\ 

Here is an example of a Semantic Triples and its corresponding reconstructed text: \newline

1. A few people | are in | a restaurant setting \newline
2. One person | is drinking | orange juice  \newline
Output format: \newline
A few people in a restaurant setting, one of them is drinking orange juice. \newline

Here is another example of a Semantic Triples and its corresponding reconstructed text: \newline
2. I | am | a student \newline
1. I | am | a professor \newline
Output format: \newline
I am a student and also a professor. \newline

Please provide no answers other than the reconstructed text. Output only the reconstructed text. And don't consider the number of sentences in the input text. \newline

Please follow the order of the inputs strictly as they are written. Do not consider the numbers provided in the inputs. For example: \newline
2. I | am | a student \newline
1. I | am | a professor \newline
Output format: \newline
I am a student and also a professor. \newline
In this case, even though the sequence numbered ``2" comes first numerically, ignore the numbers and generate the output starting with "I | am | a student" as shown in the example. \newline

Here is a Semantic Triples you should make a text: \newline
\textcolor{gray}{\# Content}
}
\vspace{1mm}
\hrule

\end{minipage}

\vfill

\newpage

\subsection{Prompt for Semantic Triple Decomposition (Natural Language Inference)}
 
\begin{minipage}[t]
{\linewidth}\raggedright
\setlength{\parindent}{0cm}
\hrule
\vspace{1mm}

\small{
\texttt{\textcolor{blue}{System:}} You are a chatbot used for data augmentation. I will provide two paragraphs or internet comments for natural language understanding tasks. This natural language understanding task has a label of entailment, contradiction, or neutral. 

\texttt{\textcolor{blue}{User:}} You should creating semantic triples from the following paragraph, and select the most important semantic triples. Your task is to receive two sentences along with the label for a natural language understanding task corresponding to those sentences. For each sentence, you need to create semantic triples. 

Here is an example of two input sentence and label: \newline

sent1: A woman is walking across the street eating a banana, while a man is following with his briefcase. \newline
sent2: An actress and her favorite assistant talk a walk in the city. \newline
label: neutral \newline

Here is an output example of semantic triples: \newline

sent1: \newline
1-1. A woman | is walking | across the street \newline
1-2. A woman | is eating | a banana \newline
1-3. A man | is following | a woman \newline
1-4. A man | is carrying | a briefcase \newline

sent2: \newline
2-1. An actress | is walking | in the city \newline
2-2. An actress | is with | her favorite assistant \newline
2-3. An actress and her favorite assistant | are talking | while walking \newline

1. A few people | are in | a restaurant setting \\
2. One person | is drinking | orange juice \newline

Here is an another example of two input sentence and label: \newline

sent1: Two women, holding food carryout containers, hug. \newline
sent2: Two groups of rival gang members flipped each other off. \newline
label: contradiction \newline

Here is an output example of above example: \newline

sent1: \newline
1-1. Two women | are holding | food carryout containers \newline
1-2. Two women | hug | each other \newline

sent2: \newline
2-1. Two groups of rival gang members | flipped | each other off \newline

Please provide no answers other than the semantic triplet. Output only the semantic triples. \newline

Here is a paragraph you should make a semantic triplet: \newline
\textcolor{gray}{\# Content}
}
\vspace{1mm}
\hrule
\end{minipage}

\vfill

\subsection{Prompt for Reconstructing Triples into Sentences (Natural Language Inference)}

\begin{minipage}[t]
{\linewidth}\raggedright
\setlength{\parindent}{0cm}
\hrule
\vspace{1mm}

\small{
\texttt{\textcolor{blue}{System:}} You are a chatbot used for data augmentation. I will provide triples for natural language understanding tasks. This natural language understanding task has a label of entailment, contradiction, or neutral. 

\texttt{\textcolor{blue}{User:}} You should reconstruct the semantic triples into a sentence or paragraph. Don't change other triplet. Then reconstruct the semantic triples into a sentence or paragraph.

Here is an example of two input triples and label: \newline

sent1: \newline
1-1. An older woman | sits | at a small table \newline
1-2. An older woman | has | orange juice \newline
1-3. Employees | are smiling | in the background \newline
1-4. Employees | are wearing | bright colored shirts \newline

sent2: \newline
2-1. A girl | flips | a burger \newline

label: contradiction \newline

Here is example of output: \newline

reconstructed sent1: \newline
An older woman sits at a small table with a glass of orange juice, while employees in bright-colored shirts smile in the background. \newline
reconstructed sent2: \newline
A girl flips a burger. \newline

Here is another example of two input triples and label: \newline

sent1: \newline
1-1. The school | is having | a special event \newline
1-2. The special event | is to show | American culture \newline
1-3. American culture | deals with | other cultures in parties \newline

sent2: \newline
2-1. A school | is hosting | an event \newline

Here is example of output: \newline

reconstructed sent1: \newline
The school is having a special event in order to show the american culture on how other cultures are dealt with in parties. \newline
reconstructed sent2: \newline
A school is hosting an event. \newline

Please follow the example format exactly and only output the necessary graph triplets. Do not start with conversational phrases like ``Here's" or ``Sure." \newline

Here is an semantic triples you should reconstruct: \newline
\textcolor{gray}{\# Content}
}
\vspace{1mm}
\hrule
\end{minipage}

\end{document}